\documentclass{article}

\usepackage{microtype}
\usepackage{graphicx}
\usepackage{stfloats}
\usepackage{subfigure}
\usepackage{amsmath}
\usepackage{booktabs} 
\usepackage{dsfont}
\usepackage{multirow}
\newcommand{\etal}{\textit{et al}.}
\newcommand{\ie}{\textit{i}.\textit{e}.}

\newcommand{\tabincell}[2]{\begin{tabular}{@{}#1@{}}#2\end{tabular}}

\usepackage{hyperref}

\usepackage[accepted]{icml2019}
\icmltitlerunning{PointCutMix}

\begin{document}

\twocolumn[
\icmltitle{PointCutMix: Regularization Strategy for Point Cloud Classification}

\icmlsetsymbol{equal}{*}

\begin{icmlauthorlist}
\icmlauthor{Jinlai Zhang}{gxu}{*}
\icmlauthor{Lyujie Chen}{thu}{*}
\icmlauthor{Bo Ouyang}{thu}
\icmlauthor{Binbin Liu}{thu}
\icmlauthor{Jihong Zhu}{gxu,thu,thu1}
\icmlauthor{Yujin Chen}{gxu}
\icmlauthor{Yanmei Meng}{gxu}
\icmlauthor{Danfeng Wu}{thu1}
\end{icmlauthorlist}

\icmlaffiliation{thu}{Department of Computer Science and Technology, Tsinghua University, Beijing, China}
\icmlaffiliation{thu1}{Department of Precision Instrument, Tsinghua University, Beijing, China}
\icmlaffiliation{gxu}{College of Mechanical Engineering, Guangxi University, Nanning, China}

\icmlcorrespondingauthor{Jihong Zhu}{jhzhu@mail.tsinghua.edu.cn}

\icmlkeywords{Machine Learning, ICML}

\vskip 0.3in
]



\printAffiliationsAndNotice{\icmlEqualContribution} 

\begin{abstract}
As 3D point cloud analysis has received increasing attention, the insufficient scale of point cloud datasets and the weak generalization ability of networks become prominent. In this paper, we propose a simple and effective augmentation method for the point cloud data, named PointCutMix, to alleviate those problems. It finds the optimal assignment between two point clouds and generates new training data by replacing the points in one sample with their optimal assigned pairs. Two replacement strategies are proposed to adapt to the accuracy or robustness requirement for different tasks, one of which is to randomly select all replacing points while the other one is to select k nearest neighbors of a single random point. Both strategies consistently and significantly improve the performance of various models on point cloud classification problems. By introducing the saliency maps to guide the selection of replacing points, the performance further improves. Moreover, PointCutMix is validated to enhance the model robustness against the point attack. It is worth noting that when using as a defense method, our method outperforms the state-of-the-art defense algorithms. 
The code is available at: {\footnotesize{\url{https://github.com/cuge1995/PointCutMix}}}.
\end{abstract}

\section{Introduction}
\label{submission}
With the rapid development of autonomous driving and robotics industries, making machines understand the real three-dimensional world has become a guarantee for safe and efficient task execution~\cite{deepplreview}. As a commonly used format for 3D data representation that can be directly obtained by Light Detection And Ranging (LiDAR) sensors, the point cloud has been widely applied in many computer vision fields~\cite{pointpillars,fastpointrcnn,pointGLR}, such as 3D object detection~\cite{pvrcnn,deformablepvrcnn}, point cloud segmentation~\cite{closer3d}, and point cloud classification~\cite{pointnet++, rscnn, dgcnn}. Following the pioneering work of PointNet~\cite{pointnet}, a series of deep-learning-based methods brought the performance of these tasks to a higher level. However, due to the complexity and costs of fine-grained 3d point cloud annotations~\cite{weaklypoints}, the scale of existing point cloud datasets is much smaller than that of the image datasets~\cite{pointmixup}, resulting in the overfitting and poor generalization of these methods~\cite{self_review}. Although researchers have explored several data augmentation techniques for point cloud analysis, such as rotation, scaling, and jittering~\cite{pointasnl,closer3d}, these kinds of augmentations ignore the shape complexity of the samples~\cite{pointaugment}, thus lead to insufficient training.

Over the past few years, mixed sample data augmentation (MSDA) for images has attached increasing interest which aims to create new training data by mixing the original training samples according to some rules~\cite{fmix,mixup_local}. There are two mainstream methods in MSDA. The first one is MixUp~\cite{mixup}, which interpolates between training samples by performing weighting on the whole image and its label. Another method is CutMix~\cite{cutmix}. It inserts a rectangle region from one image into another and then performs weighting on the image and its label by the ratio of the region size. In comparison, CutMix achieves better results across various models and datasets in image classification, weakly supervised object localization, and transfer learning to object detection.

In this paper, inspired by the success of MSDA in the image domain, we propose an MSDA strategy to the point cloud data, named PointCutMix. To adapt to its unordered feature, we first calculate the optimal assignment of two point clouds refer to MSN~\cite{MSN}. Then, two PointCutMix methods that replace the points in one sample with their optimal assigned pairs in another sample are formulated, \ie, PointCutMix-R and PointCutMix-K. The former randomly selects all replacing points while the latter selects k nearest neighbors of one random chosen point.
Experimental results demonstrate that both methods achieve consistent and significant improvements in object-level classification task on ModelNet40~\cite{modelnet40} and ModelNet10~\cite{modelnet40} datasets. We also at the first time exploit PointCutMix for point-wise classification, \ie, point cloud segmentation task. It is observed that PointCutMix can evidently improve the recognition accuracy for the uncommon categories. 
Inspired by the successful use of attention maps in CutMix~\cite{walawalkar2020attentive}, we further introduce the saliency maps to guide the selection of replacing points which achieves better results.
Additionally, we validate that PointCutMix can enhance the robustness of different models under the point cloud attack. When using as a defense method, under the point dropping attack~\cite{pointcloudsaliencymaps}, our PointCutMix-ModelNet40 pre-trained models surpass the state-of-the-art defense algorithm IF-Defense~\cite{ifdefense} by a large margin without using any transformation on the adversarial point clouds. We also perform defense to other point cloud attacks and achieve promising results.
Extensive experiments verify the effectiveness of our method. We believe this simple regularization strategy could be applied to various tasks and help future research in the 3D computer vision community.

\begin{figure*}[t]
\centering
\vskip 0.2in
\centerline{\includegraphics[width=\linewidth]{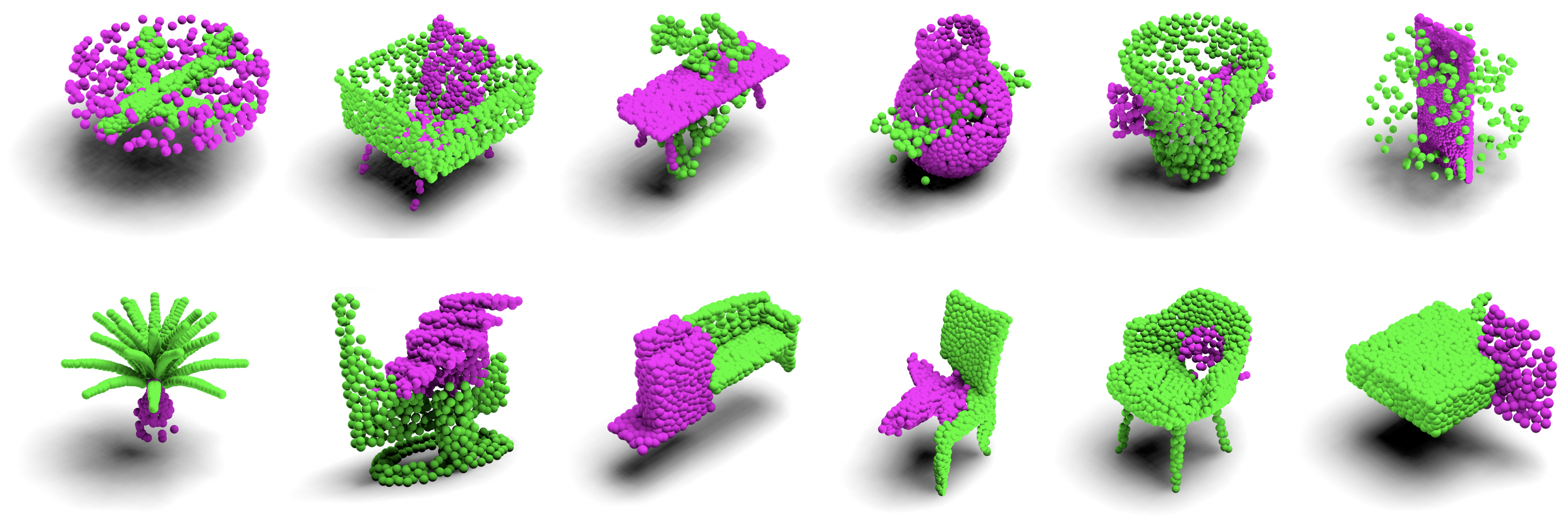}}
\caption{Some mixed samples produced by PointCutMix-R (top row) and PointCutMix-K (bottom row). The data generated by PointCutMix-R looks like two objects cross each other while the samples from PointCutMix-K are the obvious combination of two object parts.}
\label{sample}
\vskip -0.2in
\end{figure*}

\section{Related Work}

\textbf{Deep learning on point cloud.}
PointNet~\cite{pointnet} is the first work that processes the point cloud using deep neural networks where the shared pointwise multi-layer perceptions (MLPs) followed by the max-pooling operation are used for point cloud learning. After that, the recent works focus on efficiently capturing local features~\cite{pointnet++,pointasnl,pointweb,modelingpl} and investigating convolutional kernels for 3D point clouds. 
Liu~\etal~\cite{rscnn} proposed RS-CNN which implemented the convolution using an MLP in the local subset of points. DensePoint~\cite{densepoint} defined a single-layer perceptron with a nonlinear activator as convolution. In KPConv~\cite{kpconv}, by using a set of learnable kernel points, the rigid and deformable Kernel point convolution operators were proposed. 
Some other researchers have explored graph-based networks, where each point in a point cloud is considered as a vertex of a graph. DGCNN~\cite{dgcnn} constructed a graph in the feature space and MLP is used for each edge. To simplify the process of points agglomeration, the Dynamic Points Agglomeration Module based on graph convolution was proposed by Liu~\etal~\cite{dynamicPM}. In RGCNN~\cite{rgcnn}, a graph was constructed by connecting all points with each other in the point cloud. To utilize the local structural information, LocalSpecGCN~\cite{localgcn} used the spectral convolution network to a local graph.


\textbf{Mixed sample data augmentation.} 
Mixed Sample Data Augmentation (MSDA) is a strategy that produces new training data by mixing samples according to some rules~\cite{fmix}. Training with the mixed data, the model would learn multiple features in a balanced way~\cite{point_mask} and achieve better performance. Therefore, MSDA has become the mainstream data augmentation approach and dominated modern image classification for years~\cite{fmix,mixup_local,mixup,cutmix,manifold}.
Among them, MixUp~\cite{mixup} and Cutmix~\cite{cutmix} are two classical methods that have been widely used in various computer vision research~\cite{bagcnn} and competition~\cite{deepfake}.
MixUp~\cite{mixup} interpolates the training samples by performing weighting on the whole image and its label.
CutMix~\cite{cutmix} inserts a rectangle region from one image into another one and then performing weighting on the image and its label by the ratio of the region size.
The experiment results show that CutMix has better performance improvement across different datasets and networks. 
Our work can be viewed as an extension of CutMix~\cite{cutmix} for the point cloud.

\textbf{Data augmentation on point cloud.} 
Although random rotation, jittering, and scaling are commonly used in point cloud learning~\cite{pointnet,pointnet++}, the data augmentation for point clouds has obviously not been studied systematically compared to the image domain. 
Recently, PointAugment~\cite{pointaugment} and PointMixup~\cite{pointmixup} were proposed for point cloud data augmentation. PointAugment is the first auto-augmentation framework for the point cloud which optimizes the augmentor and classifier networks jointly. However, the additional augmentor network and the complicated adversarial training process makes it less practical. PointMixup extents Mixup~\cite{mixup} to point cloud by interpolation between point cloud samples. However, for point cloud networks like PointNet++~\cite{pointnet++} and RS-CNN~\cite{rscnn} that local features are important for point cloud learning, this approach is easy to fall into locally ambiguous and unnatural. In this paper, we proposed PointCutMix to naturally combine two point clouds.

\section{PointCutMix}

\subsection{Problem setting}

The goal of a standard point cloud classification task is to learn a function $f:x\rightarrow {\left [ 0,1 \right ]}^C$ that maps a point cloud to a one-hot class label for a total of $C$ classes. Here $ {x} \in \mathbf{R}^{N \times d} $ represents a set of 3D points $\{P_i | i = 1, ..., N\}$ which either sampled from a shape or pre-segmented from a scene point cloud. $N$ is the point number and each point $P_i$ is a vector with $d$ channels. In this paper, we simplicit use the 3d coordinate features. So $d=3$ and $P_i \in \mathbf{R}^{3}$. 
The optimal parameters $\theta$ of function $f$ can be learned by minimizing the loss as
\begin{equation}
\theta^{*} = \underset{\theta}{\arg \min} \sum_{x \in \mathcal{D}}^{} \mathcal{L}_{\mathcal{D}}(f(x),y)
\end{equation}
where $f(x)$ is the network output, $y$ is the ground truth with respect to $x$, $\mathcal{D}$ is the training set, and $\mathcal{L}$ represents the training loss function.


\subsection{Optimal assignment of point clouds.} 
To perform MSDA, it requires a one-to-one correspondence between the minimal unit of two samples. For image, this unit is pixel while for point cloud data, that is a single point. In the image domain, the pixels are arranged in a grid form. By merely resizing or cropping two images to the same size, it is natural to make them correspond according to their coordinate. However, the point clouds are permutation-invariant and orderless. It is essential to define the one-to-one correspondence between points based on rules other than position.

Following the method in PointMixup~\cite{pointmixup} and MSN~\cite{MSN}, we define the optimal assignment $\phi^{*}$ between two point clouds $x_1$, $x_2$ as the optimal assignment of Earth Mover's Distance (EMD)~\cite{emd} function.
The EMD calculates the minimum total displacement required for matching each point in $x_1$ to the corresponding point in $x_2$. We define the assignment function in the EMD as: 

\begin{equation}
\phi^{*}=\underset{\phi \in \Phi}{\arg \min } \sum_{i}\left\|x_{1,i}-x_{2,\phi(i)}\right\|_{2}
\end{equation}
where $
\boldsymbol{\Phi}=\{\{1, \ldots, N\} \mapsto\{1, \ldots, N\}\}
$ give one-to-one correspondences between the two point clouds. After given the optimal assignment $\phi^{*}$~\cite{pointmixup}, the EMD is then defined as:
\begin{equation}
\mathrm{EMD}=\frac{1}{N} \sum_{i}\left\|x_{1,i}-x_{2,\phi^{*}(i)}\right\|_{2}
\end{equation}
where $\phi^{*}(i)$ denotes the index of optimal assignment point of $x_{1,i}$ in $x_{2}$.

\subsection{Algorithm}
The key idea of PointCutMix is to create a new training point cloud $
(\tilde{x}, \tilde{y})$ given two distinct training point clouds $
({x_1}, {y_1})$ and $({x_2}, {y_2})$. Here, $x$ is the training point cloud and $y$ is the corresponding label. After obtaining the optimal assignment $\phi^{*}$ between two samples, we define $\tilde{x_{2,i}}=x_{2,\phi^{*}(i)}$ and the combining operation as

\begin{equation}
\begin{array}{l}
\tilde{x}=B \cdot x_{1}+(I_N-B) \cdot \tilde{x_{2}} \\ 
\tilde{y}=\lambda y_{1}+(1-\lambda) y_{2}
\end{array}
\end{equation}

where $B=diag\{ b_1, b_2, \cdots, b_N \}$ and $b_i \in \{0,1\}$ indicates which sample the point belongs to. When $b_i=1$, the $i^{th}$ point is chosen from $x_1$, otherwise it will replaced by the optimal assigned point in $x_2$.
$I_N$ is an identity matrix. $\lambda \in [0,1]$ is the PointCutMix ratio, sampled from the beta distribution $Beta(\beta, \beta)$, which means $n=\left \lfloor \lambda \times N \right \rfloor$ points will be kept and the rest points will be replaced.

To perform cutting and pasting in the point cloud, we propose two replacement methods to construct the diagonal matrix $B$.
The first method, abbreviated as PointCutMix-R, is to randomly sample $n$ points from $x_{1}$ as a subset $x_{1}^s$. Those points are marked $1$ in $B$, indicating that they will not be replaced. The rest points are marked $0$.
In addition, to retain the local characteristics of the point cloud, we come up with the second method, noted as PointCutMix-K, which randomly sample one central point $p$ from $x_{1}$, and then finding its $n-1$ nearest neighbors. We combine $p$ and its nearest neighbors to form $x_{1}^s$ and marked those points as $1$ in $B$. Similarly, the rest points are marked $0$ and be replaced.
In Figure~\ref{sample}, we show the visualization of some mixed samples of PointCutMix-R and PointCutMix-K. It can be seen that the samples produced by PointCutMix-R look like two objects cross together while the mixed data from PointCutMix-K are the obvious combination of two object parts.

We also introduce a hyperparameter $\rho \in \left [ 0,1 \right ]$ to indicate the probability of each point cloud to be augmented during the training. When $\rho=0$, PointCutMix will not be used which is equivalent to the baseline model. On the contrary, $\rho=1$ means all point clouds will be augmented. Therefore, the training loss can be denoted as

\begin{figure*}[t]
\vskip 0.2in
\begin{center}
\centerline{\includegraphics[width=\linewidth]{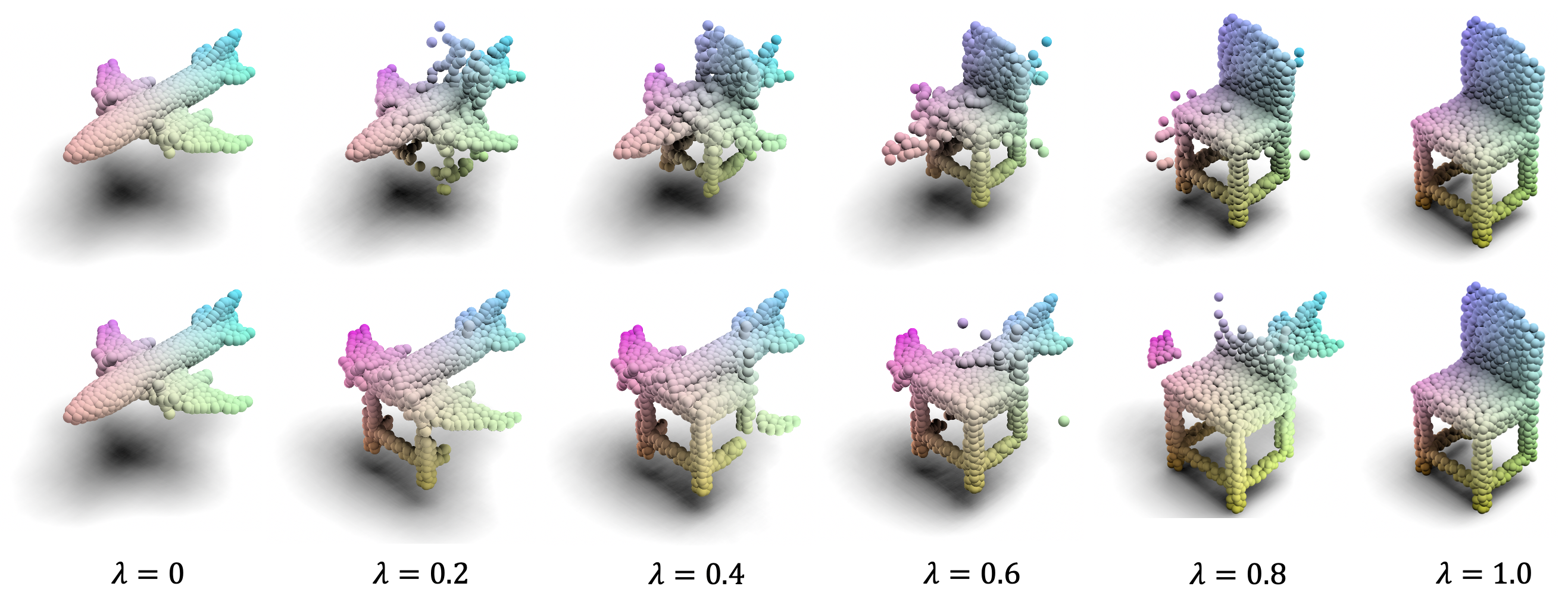}}
\caption{The visualization of the mixed samples between a plane and a chair under different replacement ratio $\lambda$. The samples in the first and second row are generated by PointCutMix-R and PointCutMix-K respectively.}
\label{ratio}
\end{center}
 \vskip -0.2in
\end{figure*}

\begin{equation}
\sum_{x \in \mathcal{D}}^{} (1-\mathds{1}_\rho) \mathcal{L}_{\mathcal{D}}(f(x),y) + \mathds{1}_\rho \mathcal{L}_{\mathcal{D}}(f(\tilde{x}),\tilde{y})
\end{equation}

where $\mathds{1}_\rho = 1$ with a probability $\rho$, otherwise it equals to 0. 

\subsection{Analysis}

\begin{figure}[t]
\vskip 0.2in
\begin{center}
\centerline{\includegraphics[width=\linewidth]{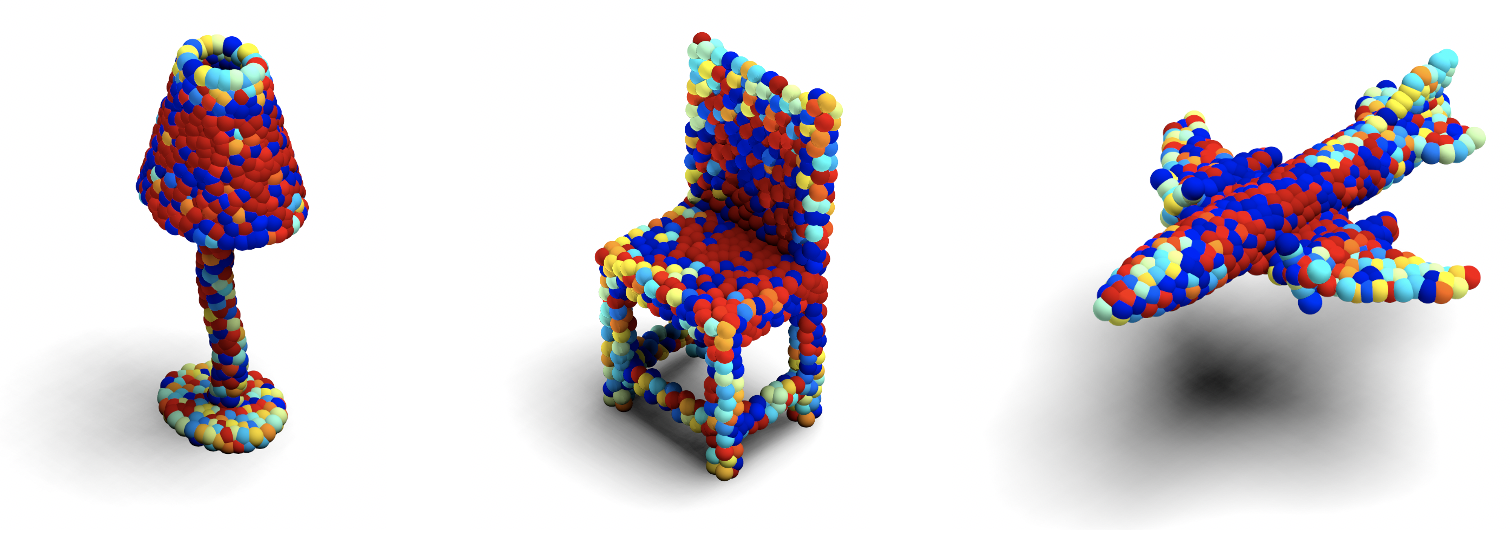}}
\caption{Saliency maps of different point clouds. Points with higher values are colored as red and the color of irrelevant points is closer to blue.}
\label{cam_smap}
\end{center}
\vskip -0.2in
\end{figure}

\begin{figure}[t]
\vskip 0.2in
\begin{center}
\centerline{\includegraphics[width=\linewidth]{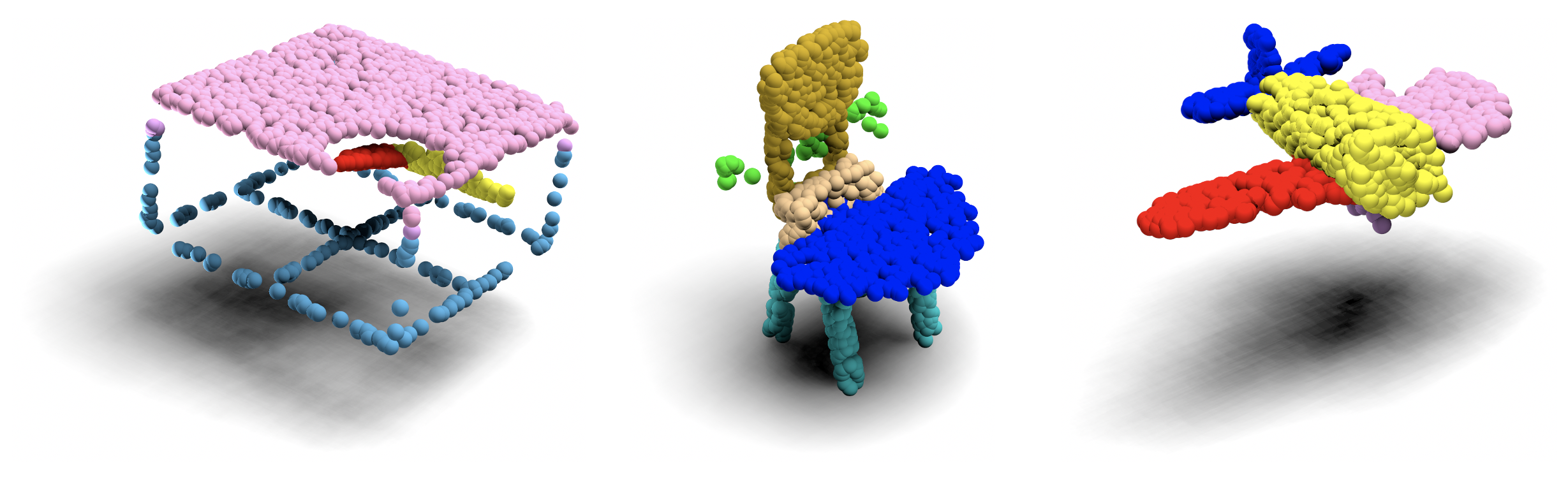}}
\caption{Mixed samples of point cloud segmentation problem using PointCutMix-K.}
\label{seg}
\end{center}
\vskip -0.2in
\end{figure}

\textbf{The difference of replacement methods.} 
In Figure~\ref{ratio}, we list the visualization of mixed samples between a plane and a chair produced by PointCutMix-R (top row) and PointCutMix-K (bottom row) under different replacement ratios $\lambda$. The ratios from left to right are 0, 0.2, 0.4, 0.6, 0.8, and 1.0. 
We can see that the samples in the top row are a little messy and like two objects fuse. Especially when $\lambda$ close to 0 or 1, such as $\lambda=0.2$ or $\lambda=0.8$, only one of the objects can be easily recognized. Those hard to distinguished points actually perform like a kind of noise. We infer that this characteristic will impair the performance for learning classification task, but can improve the robustness of the model. This assumption has been verified in the experiment in Section~4.
On the contrary, the mixed samples from PointCutMix-K are relatively regular, like a natural combination of two object parts. Since at least a part of each object can be identified, it provides more features for learning the classification task.

\textbf{Is attention works for PointCutMix?}
Inspired by the successful use of attention maps to guide the cutting and pasting region of an object in CutMix~\cite{walawalkar2020attentive}, we speculate that the selection of central point $p$ in PointCutMix-K can also be guided rather than random. 
So we try to obtain the contribution of each point to the classification result with saliency map~\cite{pointcloudsaliencymaps}. 
Then the point with greater contribution has a higher probability to be selected as the central point. Through the visualization of the saliency maps shown in Figure~\ref{cam_smap}, we find that the points with high contributions (in red color) are not scattered uniformly. For example, they gather more in the lampshades, the seat of the chair, and the fuselage of the airplane. In the next section, we will examine whether this strategy improves the accuracy of the model.

\textbf{Extending to point cloud segmentation.} 
So far, we have elaborated on how to apply our method to the object-level classification. Natural intuition is to extend to the point-wise classification problem, \ie, point cloud segmentation. However, we find that the existing augmentation methods are limited by their fusion strategies, thus fail to complete this task. For example, when applying PointMixup to fuse a new point cloud, the semantic information of each point has lost, making it hard to get the semantic labels for new data. 
On the contrary, since our method is simply cutting and pasting points, the semantic information can be persisted. So in this paper, we at the first time perform augmentation to point cloud segmentation task. Specifically, we mix two point clouds using the same method mentioned before. The point-wise labels are replaced along with the points. For datasets that also contain object-level labels, the object-level annotation is fused referred to Section 3.3. In Figure~\ref{seg}, we show some mixed samples for the point cloud segmentation problem.





\section{Experiments}

In this section, we conduct extensive experiments to verify the effectiveness of PointCutMix. At first, we find the optimal hyperparameters through several comparative experiments. Then we assess our method from two aspects, one of which is to evaluate how much it improves the accuracy of object-level point cloud classification and point-wise segmentation while the other one is to evaluate the generalization ability and robustness of the model trained with augmented data provided by PointCutMix.

\subsection{Setup}

\textbf{Datasets.} 
We evaluate PointCutMix on two object-level point cloud classification datasets and a point-wise segmentation dataset, \ie, ModelNet40~\cite{modelnet40}, ModelNet10~\cite{modelnet40}, and ShapeNet Parts~\cite{shapepartseg}. 
ModelNet40 contains 12311 samples in 40 categories. Among them, 9843 samples are used for training and 2468 for testing. 
ModelNet10 is a subset of ModelNet40. It contains a total of 4899 samples in 10 categories, of which 3991 samples are used for training and the rest are used for testing.
ShapeNet Parts consists of 16,880 3D samples in 16 categories and 50 part labels, of which 14,006 for training and 2,874 for testing.

\textbf{Networks.} Since PointCutMix is a general data augmentation method, it is agnostic to the network architecture that is employed. Therefore, we select four popular networks in 3D computer vision area~\cite{pointaugment,isometry3d} for evaluation, \ie, PointNet~\cite{pointnet}, PointNet++~\cite{pointnet++}, RS-CNN~\cite{rscnn}, and DGCNN~\cite{dgcnn}. As mentioned in Section 2, PointNet only uses global information while other three models take the local information into account.

\textbf{Implementation details.} Our work is implemented using PyTorch~\cite{torch} on NVIDIA GeForce GTX 2080Ti GPU. All networks take 1024 points as input and are trained for 300 epochs with a batch size of 16. For PointNet, PointNet++, and RS-CNN, we use the Adam~\cite{kingma2014adam} optimizer with an initial learning rate of 0.001 and a decay rate of 0.5 every 20 epochs, which is the same configuration as the original released paper and code. We train DGCNN with SGD optimizer with an initial learning rate of 0.1. The minimum learning rate is 0.001 and the momentum of SGD is 0.9. The cosine annealing strategy is used to decay the learning rate.


\begin{figure}[t]
\vskip 0.2in
\begin{center}
\centerline{\includegraphics[width=0.8\linewidth]{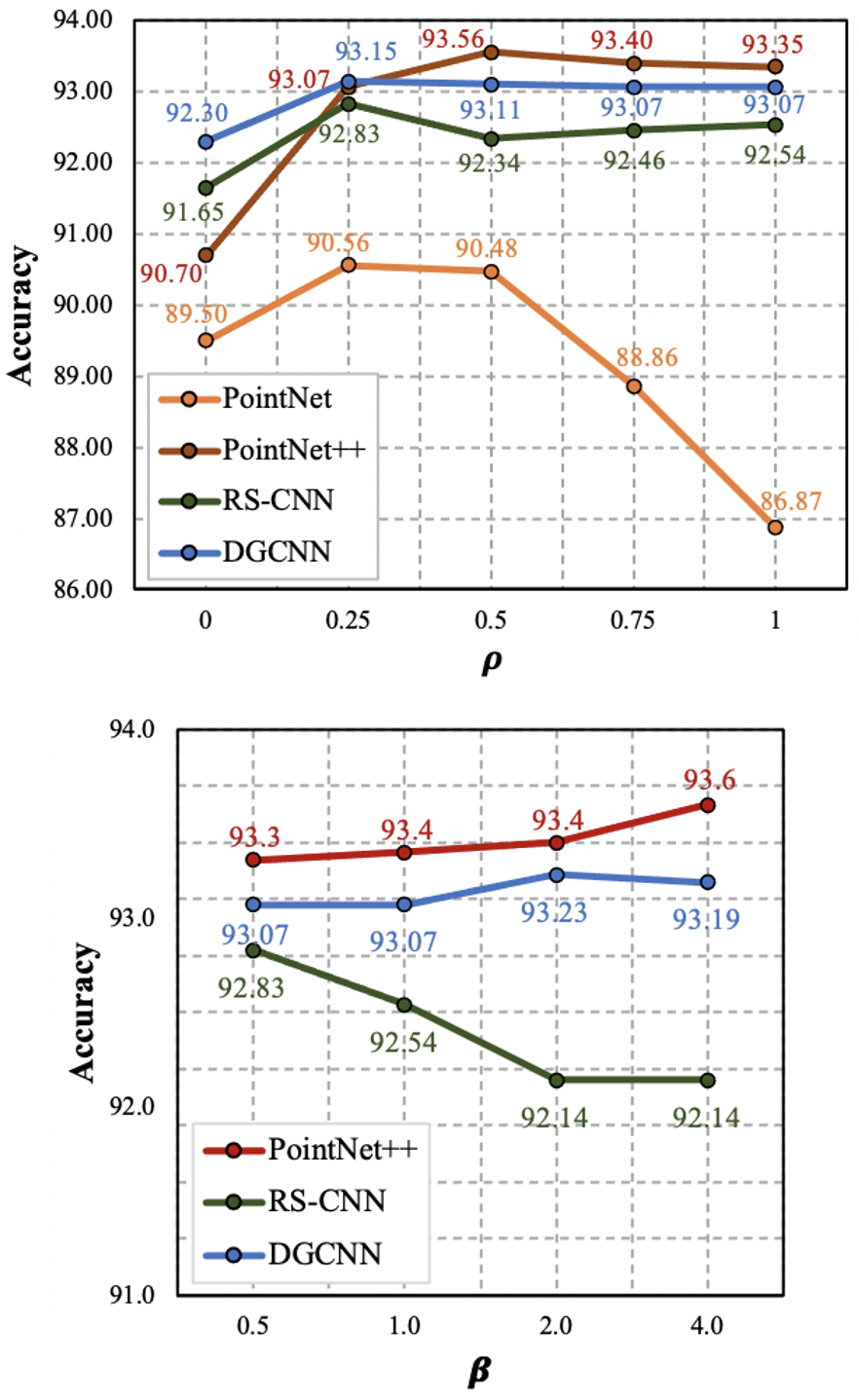}}
\caption{Performance of various models with PointCutMix-K under different value of $\rho$ and $\beta$. In the upper plot, $\beta=1$. In the lower plot, $\rho=1$.}
\label{prob_beta}
\end{center}
\vskip -0.2in
\end{figure}

\subsection{Comparative experiments}
We perform the comparative experiments in ModelNet40 dataset using the experimental settings described in implementation details.

\textbf{Influence of $\rho$.} 
We first compare the performance of PointCutMix-K on four representative models under different values of $\rho$ to figure out whether our method is useful and how much of the data need to be augmented during the training. The results are illustrated in Figure~\ref{prob_beta}.
For pointNet++, RS-CNN, and DGCNN, we observe that even with only 25\% of the samples are augmented ($\rho=0.25$), the accuracy is greatly improved, which proves that our method is very essential and effective. Under different values of $\rho$, there is no much difference in accuracy for three models.
However, PointNet performs in a completely different way. It is improved when $\rho$ is small, but the accuracy is significantly dropped when $\rho$ reaches 1. We speculate that this is because the coordinate feature of a single point has no actual information. The object-level classification must rely on the learning of the relationship between points. However, PointNet lacks the ability to learn local features since it performs MLP for all points in the object together, which makes it difficult to distinguish the replaced region. 

In the following experiments, although each model reaches its optimal performance with different values of $\rho$, we choose $\rho=1.0$ for the object-level point cloud classification task in order to make a fair comparison to other augmentation methods. Here we do not report PointNet since our method is not suitable for it. While for the point cloud segmentation task, we select $\rho=0.5$ for better performance.

\begin{table}[t]
\centering
\setlength{\tabcolsep}{1.8mm}
\caption{ModelNet40 classification results. PointMixup-U and PointMixup-A represent the results on unaligned and pre-aligned ModelNet40 with input mixup.}
\label{mn40result}
\vspace{1mm}
\begin{tabular}{l c c c}
\toprule
Method & PointNet++ & RS-CNN & DGCNN \\
\midrule
baseline & 90.7 & 91.7 & 92.3 
\\
PointMixup-U & 91.7 & - & -
\\
PointMixup-A & 92.7 & - & 92.9
\\
PointAugment & 92.9 & 92.7 & \textcolor{blue}{\textbf{93.4}}
\\
\midrule
PointCutMix-R & 92.8 & 91.9 & 92.8
\\
PointCutMix-K & 93.4 & 92.5 & 93.1
\\
PointCutMix-S & \textcolor{blue}{\textbf{93.4}} & \textcolor{blue}{\textbf{92.7}} & 93.2
\\
\bottomrule
\end{tabular}
\end{table}

\begin{table}[tp]
\centering
\setlength{\tabcolsep}{1.8mm}
\caption{ModelNet10 classification results.}
\label{mn10result}
\vspace{1mm}
\begin{tabular}{l c c c}
\toprule
Method & Pointnet++ & RS-CNN & DGCNN \\
\midrule
baseline &93.3 & 94.2 & 94.8
\\
PointAugment & 95.8 & \textcolor{blue}{\textbf{96.0}} & \textcolor{blue}{\textbf{96.7}}
\\
\midrule
PointCutMix-R & \textcolor{blue}{\textbf{96.3}}  & 95.7  &  95.2
\\
PointCutMix-K & 95.7 & 95.6 & 95.7
\\
\bottomrule
\end{tabular}
\end{table}




\textbf{Influence of $\beta$.} Next, we investigate the influence of $\beta$, \ie, whether there is a difference in choosing the different number of replaced points at augmentation. From the results in Figure~\ref{prob_beta}, it can be seen that the difference of accuracy under various values of $\beta$ for PointNet++ and DGCNN is very small, but RSCNN prefers a small value of $\beta$. To use the same hyperparameter for all models and simplify the selecting process of $\alpha$, we select the $Beta(1,1)$, \ie, the uniform distribution in the subsequent experiments.

\subsection{Point cloud classification}

After determining the hyperparameters, we conduct point cloud classification experiments on ModelNet40 and ModelNet10 to evaluate various data augmentation methods, including conventional data augmentation (baseline)~\cite{pointnet++}, PointMixup~\cite{pointmixup}, PointAugment~\cite{pointaugment}, PointCutMix-R, and PointCutMix-K. In addition, to verify the influence of attention maps mentioned in Section~3.4, we introduce the saliency map to guide the selection of central point $p$. This strategy is named PointCutMix-S.
The results of baseline models refer to PointAugment. The models trained with PointCutMix methods are implemented with the settings in our implementation details. The saliency maps are produced by corresponding pre-trained baseline models during the training. The results of PointMixup and PointAugment refer to their original papers. 

From Table~\ref{mn40result} and Table~\ref{mn10result}, we observe that our methods consistently outperform PointMixup and have comparative results to PointAugment. This is a very impressive result because PointCutMix is much simpler than the existing methods. PointMixup needs to pre-align the point clouds of the training and test sets in the horizontal facing direction. But our method does not rely on any pre-process for the input point clouds. PointAugment uses an additional network for data augmentation. It requires much more memory cost, which is not practical in real applications. In comparison, PointCutMix uses little computing resources and time but still achieves better performance. 

PointCutMix-R occasionally has better results than PointCutMix-K on ModelNet10. However, in most cases across two datasets, PointCutMix-K performs better. The results also show that the saliency maps have limited help for the performance. Considering the addition calculation time and memory consumption used for generating the saliency maps during training, we hold that PointCutMix-K is a more versatile and efficient strategy.

\begin{table*}[t!]
\centering
\caption{Comparison on the ShaperNet part segmentation dataset. pIoU means part-average Intersection-over-Union. We perform the experiment using the settings described in Section 4.4.}
\setlength{\tabcolsep}{1.9pt}
\renewcommand\arraystretch{1.15}
\begin{tabular}{c|c|cccccccccccccccc}
\hline
\textbf{Method} & pIoU& \tabincell{c}{{air}-\\{plane}} & bag & cap& car& chair&\tabincell{c}{{ear}-\\{phone}} & guitar& {knife}& {lamp}& {laptop}&\tabincell{c}{motor-\\bike} & {mug} & {pistol} & {rocket} & \tabincell{c}{{skate}-\\{board}}& {table} \\
\hline

PointNet++     & 85.0          & 82.2                                                 & 81.7          & 81.5          & 77.7          & 90.1          & 76.7                                                 & 90.9          & 87.3          & 83.8          & 95.2          & 69.9                 & 94.2          & 82.6          & 56.2          & 76.6          & 82.8          \\
+PointCutMix & 85.5 & 82.6 & \textbf{85.9} & \textbf{83.7} & 78.3          & 90.7          & 72.5  & 90.9          & 87.7          & 84.3          & 95.3          & 70.7                 & 95.1          & 82.4          & \textbf{62.3}          & 74.9          & 83.4          \\
        \hline
    \end{tabular}
    \label{seg_result}
    \vspace{-1ex}
\end{table*}

\subsection{Point cloud segmentation}
To explore the extensibility of our method, we at the first time apply augmentation to the point cloud segmentation task. Here we train the baseline model and PointCutMix-S for 251 epochs with a batch size of 16. We use Adam~\cite{kingma2014adam} optimizer with an initial learning rate of 0.001 and a decay rate of 0.5 every 20 epochs. In Table~\ref{seg_result}, we report the part-average Intersection-over-Union results. It shows that PointCutMix makes an improvement of 0.5\% over the PointNet++ baseline. Although the improvement is not as significant as that for the object-level classification task, there is a special finding that the accuracy gains mainly come from the uncommon categories. Specifically, the ShapeNet Parts dataset~\cite{shapepartseg} has an uneven distribution of training data, where the table has 5271 samples but the bag, cap, and rocket have only 76, 55, and 66 samples respectively. Training with our PointCutMix augmentation method, over 6.1 pIoU improvement is made for the rocket part-segmentation.

We infer the reason is that through the fusion of training samples in PointCutMix, the frequency of occurrence of uncommon categories is greatly increased. This also enlightens us that by carefully adjusting the ratio of selecting different categories of samples for augmentation, the unbalanced distribution problem might be effectively alleviated, which is worth exploring in the future.

\begin{table}[t]
\centering
\setlength\tabcolsep{3pt}
\caption{Classification accuracy of ModelNet40 under point dropping attack~\cite{pointcloudsaliencymaps} , the dropping points is 200.}
\label{attackresult}
\vspace{2mm}
\begin{tabular}{l c c c c}
\toprule
Model & Baseline  & PointCutMix-R & PointCutMix-K \\
\midrule
PointNet++ & 68.96 & 86.18 & \textcolor{blue}{\textbf{87.97}} 
\\
RS-CNN &56.97 &82.50 & \textcolor{blue}{\textbf{83.10}}
\\
DGCNN &55.06  & 81.16  & \textcolor{blue}{\textbf{85.86}}
\\
\bottomrule
\end{tabular}
\end{table}

\begin{table*}[t]
\centering
\setlength{\tabcolsep}{3pt}
\renewcommand\arraystretch{1.15}
\caption{Classification accuracy of various defense methods on ModelNet40 under point dropping attack~\cite{pointcloudsaliencymaps}, kNN attack~\cite{knnattack} and point perturbation attack~\cite{generatingadpoint}. Drop 200 and Drop 100 denote the dropping points is 200 and 100 respectively. $*$ denotes that results are reported in IF-Defense~\cite{ifdefense}. We report the best result of three IF-Defense methods. The best and second-place results for each row are emphasized as blue and bold.}
\label{drop}
\vspace{2mm}
\begin{tabular}{c c c c c c c c c}
\toprule
Attack & Model & No Defense$*$ & SRS$*$ & SOR$*$ & DUP-Net$*$ & IF-Defense$*$ & PointCutMix-R & PointCutMix-K\\
\midrule
\multirow{2}{*}{Drop 200} & 
PointNet++ & 68.96 & 39.63 & 69.17 & 72.00 & 79.09 & \textbf{87.32} & \textcolor{blue}{\textbf{89.02}}
\\
\cline{2-9}~& DGCNN & 55.06 & 23.82 & 59.36 & 36.02 & 73.30 & \textbf{87.36} & \textcolor{blue}{\textbf{88.82}}
\\
\midrule
\multirow{2}*{Drop 100} & 
PointNet++ & 80.19 & 64.51 & 74.16 & 76.38 & 84.56 & \textbf{89.51} & \textcolor{blue}{\textbf{91.17}}
\\
\cline{2-9}~ &DGCNN & 75.16 & 49.23 & 64.68 & 44.45 & 83.43 & \textbf{89.59} & \textcolor{blue}{\textbf{91.05}}
\\
\midrule
\multirow{2}{*}{kNN} &
PointNet++ & 0.00&49.96 & 61.35& 74.88&  \textcolor{blue}{\textbf{85.62}} & \textbf{83.35} &  70.71 
\\
\cline{2-9}~&DGCNN &20.02 &41.25 & 55.92& 35.45&  \textcolor{blue}{\textbf{82.33}} & \textbf{80.27} & 68.76
\\
\midrule
\multirow{2}*{point perturbation} &
PointNet++ &0.00 & 73.14 & 77.67& 80.63& \textcolor{blue}{\textbf{86.99}} & \textbf{86.71} & 84.93
\\
\cline{2-9}~ &DGCNN &0.00 & 50.20 & 76.50& 42.67& \textcolor{blue}{\textbf{85.53}} & \textbf{83.14} & 76.69
\\

\bottomrule
\end{tabular}
\end{table*}

\subsection{Robustness test}
After verifying the accuracy improvement of PointCutMix on the point cloud classification, we then use the adversarial attack to investigate whether this regularization strategy can enhance the robustness of the model.
As we know, deep neural networks are vulnerable to adversarial examples, which have been extensively studied in 2D images~\cite{boostingad,adsurvey}. Recently, point perturbation attack~\cite{generatingadpoint}, kNN attack~\cite{knnattack}, and point dropping attack~\cite{pointcloudsaliencymaps} are proposed for 3D point cloud. In this paper, our method is trained after the normalization of point clouds. Since the point perturbation attack and the kNN attack don't perform normalization of point clouds during the attack and the generated point clouds may not center within a unit sphere, we only consider the point dropping attack in our robustness test. 

We report the recognition accuracy after the point dropping attack on the test set of ModelNet40 in Table~\ref{attackresult}, where the results of baseline models refer to IF-Defense~\cite{ifdefense}. 
It is observed that the baseline model dramatically degrade. But all models trained with PointCutMix-R and PointCutMix-K still have more than 80\% accuracy. It verifies that our method can significantly improve the robustness of the model.

\subsection{Point cloud defense}
Motivated by the impressive performance under point drop attack, we consider applying our method to the point cloud defense. We surprisingly find that using the pre-trained models trained with PointCutMix augmentation as defense methods outperforms the state-of-the-art defense algorithm IF-Defense~\cite{ifdefense} by a large margin. Specifically, we first generate adversarial point clouds by point dropping attack on the pre-trained baseline model provided by \cite{ifdefense}. We then compare the classifiers trained using PointCutMix augmentation method on these generated adversarial point clouds to several recent developed defense methods, \ie, SRS~\cite{adversarialpl}, SOR~\cite{dupnet}, DUP-Net~\cite{dupnet} and IF-Defense~\cite{ifdefense}. 
From the results listed in Table~\ref{drop}, we observe that PointCutMix-R and PointCutMix-K consistently surpass all defense methods under two point dropping attacks. The improvement of recognition accuracy can reach up to 15\% in a certain case, which fully proves the scalability and effectiveness of our method.
It is worth noting that unlike previous defense methods that need to alter the adversarial point clouds which might cause information loss, our method just uses very limited computing power to classify the adversarial point clouds, which is a more natural defense approach.

Moreover, to verify the generalization of PointCutMix in defense, we also test with the kNN attack~\cite{knnattack} and the point perturbation attack~\cite{generatingadpoint}.
We first perform normalization on the generated adversarial point clouds to limit all points into a unit sphere and then test it with deep point cloud classification networks trained using PointCutMix-K and PointCutMix-R. 
The kNN attack smoothes the attack by using a k-Nearest Neighbor loss, which is hard to defense by the simple method such as statistical outlier removal~\cite{dupnet}. On the contrary, for the point perturbation attack~\cite{generatingadpoint} where the attacked point clouds are messy, it can be easily defended by simple random sampling and statistical outlier removal~\cite{dupnet}. 
As shown in Table~\ref{drop}, although these two attacks are very unfavorable for our models that are trained with normalized point clouds, PointCutMix-R still achieves second place and has a very close performance to the state-of-the-art defense method in all cases.
We can also find that PointCutMix-R constantly surpasses PointCutMix-K in the defense of two attacks, proving the assumption in Section 3.4 that models trained with PointCutMix-R achieve better robustness.

From the above analysis, we can conclude that PointCutMix has strong generalization ability across various point cloud attack algorithms and the defense approach is very simple and computing cost-effective.

\section{Conclusion}

In this paper, we propose PointCutMix, a regularization strategy for point cloud classification. We conduct extensive experiments to verify the effectiveness of our method. For the object-level point cloud classification problem, the results show that PointCutMix evidently improves the performance of networks that learned with local features. While for the point-wise segmentation task, PointCutMix alleviates the unbalanced distribution problem and enhances the performance of uncommon categories. We also validate that PointCutMix significantly enhances the robustness of the model. By applying our method as a defense method, it outperforms the SOTA defense algorithm. We hope this simple regularization strategy could be applied to more tasks and help future researches.

In the future, we plan to extend our work to 3D object detection~\cite{pvrcnn}. However, due to the point cloud is different from images, there are still some challenges. For example, in 3D object detection, the point cloud of KITTI~\cite{kitti} and ModelNet are very different, thus it is hard to directly use the pre-trained model of the classification network in the 3D detection task. Moreover, we also plan to apply our PointCutMix-R and PointCutMix-K to defense methods to recently proposed attacks AdvPC~\cite{advpc} and LG-GAN~\cite{lggan}.



\bibliography{pointcutmix}

\begin{thebibliography}{50}
\providecommand{\natexlab}[1]{#1}
\providecommand{\url}[1]{\texttt{#1}}
\expandafter\ifx\csname urlstyle\endcsname\relax
  \providecommand{\doi}[1]{doi: #1}\else
  \providecommand{\doi}{doi: \begingroup \urlstyle{rm}\Url}\fi

\bibitem[Akhtar \& Mian(2018)Akhtar and Mian]{adsurvey}
Akhtar, N. and Mian, A.
\newblock Threat of adversarial attacks on deep learning in computer vision: A
  survey.
\newblock \emph{Ieee Access}, 6:\penalty0 14410--14430, 2018.

\bibitem[Bhattacharyya \& Czarnecki(2020)Bhattacharyya and
  Czarnecki]{deformablepvrcnn}
Bhattacharyya, P. and Czarnecki, K.
\newblock Deformable pv-rcnn: Improving 3d object detection with learned
  deformations.
\newblock \emph{arXiv preprint arXiv:2008.08766}, 2020.

\bibitem[Chen et~al.(2019)Chen, Liu, Shen, and Jia]{fastpointrcnn}
Chen, Y., Liu, S., Shen, X., and Jia, J.
\newblock Fast point r-cnn.
\newblock In \emph{Proceedings of the IEEE/CVF International Conference on
  Computer Vision}, pp.\  9775--9784, 2019.

\bibitem[Chen et~al.(2020)Chen, Hu, Gavves, Mensink, Mettes, Yang, and
  Snoek]{pointmixup}
Chen, Y., Hu, V.~T., Gavves, E., Mensink, T., Mettes, P., Yang, P., and Snoek,
  C.~G.
\newblock Pointmixup: Augmentation for point clouds.
\newblock In \emph{European Conference on Computer Vision}, pp.\  330--345.
  Springer, 2020.

\bibitem[Dolhansky et~al.(2020)Dolhansky, Bitton, Pflaum, Lu, Howes, Wang, and
  Ferrer]{deepfake}
Dolhansky, B., Bitton, J., Pflaum, B., Lu, J., Howes, R., Wang, M., and Ferrer,
  C.~C.
\newblock The deepfake detection challenge dataset.
\newblock \emph{arXiv preprint arXiv:2006.07397}, 2020.

\bibitem[Dong et~al.(2018)Dong, Liao, Pang, Su, Zhu, Hu, and Li]{boostingad}
Dong, Y., Liao, F., Pang, T., Su, H., Zhu, J., Hu, X., and Li, J.
\newblock Boosting adversarial attacks with momentum.
\newblock In \emph{Proceedings of the IEEE conference on computer vision and
  pattern recognition}, pp.\  9185--9193, 2018.

\bibitem[Geiger et~al.(2012)Geiger, Lenz, and Urtasun]{kitti}
Geiger, A., Lenz, P., and Urtasun, R.
\newblock Are we ready for autonomous driving? the kitti vision benchmark
  suite.
\newblock In \emph{2012 IEEE Conference on Computer Vision and Pattern
  Recognition}, pp.\  3354--3361. IEEE, 2012.

\bibitem[Guo et~al.(2019)Guo, Mao, and Zhang]{mixup_local}
Guo, H., Mao, Y., and Zhang, R.
\newblock Mixup as locally linear out-of-manifold regularization.
\newblock In \emph{Proceedings of the AAAI Conference on Artificial
  Intelligence}, volume~33, pp.\  3714--3722, 2019.

\bibitem[Guo et~al.(2020)Guo, Wang, Hu, Liu, Liu, and Bennamoun]{deepplreview}
Guo, Y., Wang, H., Hu, Q., Liu, H., Liu, L., and Bennamoun, M.
\newblock Deep learning for 3d point clouds: A survey.
\newblock \emph{IEEE transactions on pattern analysis and machine
  intelligence}, 2020.

\bibitem[Hamdi et~al.(2020)Hamdi, Rojas, Thabet, and Ghanem]{advpc}
Hamdi, A., Rojas, S., Thabet, A., and Ghanem, B.
\newblock Advpc: Transferable adversarial perturbations on 3d point clouds.
\newblock In \emph{European Conference on Computer Vision}, pp.\  241--257.
  Springer, 2020.

\bibitem[Harris et~al.(2020)Harris, Marcu, Painter, Niranjan, and Hare]{fmix}
Harris, E., Marcu, A., Painter, M., Niranjan, M., and Hare, A. P.-B.~J.
\newblock Fmix: Enhancing mixed sample data augmentation.
\newblock \emph{arXiv preprint arXiv:2002.12047}, 2\penalty0 (3):\penalty0 4,
  2020.

\bibitem[He et~al.(2019)He, Zhang, Zhang, Zhang, Xie, and Li]{bagcnn}
He, T., Zhang, Z., Zhang, H., Zhang, Z., Xie, J., and Li, M.
\newblock Bag of tricks for image classification with convolutional neural
  networks.
\newblock In \emph{Proceedings of the IEEE/CVF Conference on Computer Vision
  and Pattern Recognition}, pp.\  558--567, 2019.

\bibitem[Jing \& Tian(2020)Jing and Tian]{self_review}
Jing, L. and Tian, Y.
\newblock Self-supervised visual feature learning with deep neural networks: A
  survey.
\newblock \emph{IEEE Transactions on Pattern Analysis and Machine
  Intelligence}, 2020.

\bibitem[Kingma \& Ba(2014)Kingma and Ba]{kingma2014adam}
Kingma, D.~P. and Ba, J.
\newblock Adam: A method for stochastic optimization.
\newblock \emph{arXiv preprint arXiv:1412.6980}, 2014.

\bibitem[Lang et~al.(2019)Lang, Vora, Caesar, Zhou, Yang, and
  Beijbom]{pointpillars}
Lang, A.~H., Vora, S., Caesar, H., Zhou, L., Yang, J., and Beijbom, O.
\newblock Pointpillars: Fast encoders for object detection from point clouds.
\newblock In \emph{Proceedings of the IEEE/CVF Conference on Computer Vision
  and Pattern Recognition}, pp.\  12697--12705, 2019.

\bibitem[Li et~al.(2020)Li, Li, Heng, and Fu]{pointaugment}
Li, R., Li, X., Heng, P.-A., and Fu, C.-W.
\newblock Pointaugment: an auto-augmentation framework for point cloud
  classification.
\newblock In \emph{Proceedings of the IEEE/CVF Conference on Computer Vision
  and Pattern Recognition}, pp.\  6378--6387, 2020.

\bibitem[Liu et~al.(2019{\natexlab{a}})Liu, Ni, Li, Yang, and Tian]{dynamicPM}
Liu, J., Ni, B., Li, C., Yang, J., and Tian, Q.
\newblock Dynamic points agglomeration for hierarchical point sets learning.
\newblock In \emph{Proceedings of the IEEE/CVF International Conference on
  Computer Vision}, pp.\  7546--7555, 2019{\natexlab{a}}.

\bibitem[Liu et~al.(2020{\natexlab{a}})Liu, Sheng, Yang, Shao, and Hu]{MSN}
Liu, M., Sheng, L., Yang, S., Shao, J., and Hu, S.-M.
\newblock Morphing and sampling network for dense point cloud completion.
\newblock In \emph{Proceedings of the AAAI Conference on Artificial
  Intelligence}, volume~34, pp.\  11596--11603, 2020{\natexlab{a}}.

\bibitem[Liu et~al.(2019{\natexlab{b}})Liu, Fan, Meng, Lu, Xiang, and
  Pan]{densepoint}
Liu, Y., Fan, B., Meng, G., Lu, J., Xiang, S., and Pan, C.
\newblock Densepoint: Learning densely contextual representation for efficient
  point cloud processing.
\newblock In \emph{Proceedings of the IEEE/CVF International Conference on
  Computer Vision}, pp.\  5239--5248, 2019{\natexlab{b}}.

\bibitem[Liu et~al.(2019{\natexlab{c}})Liu, Fan, Xiang, and Pan]{rscnn}
Liu, Y., Fan, B., Xiang, S., and Pan, C.
\newblock Relation-shape convolutional neural network for point cloud analysis.
\newblock In \emph{Proceedings of the IEEE/CVF Conference on Computer Vision
  and Pattern Recognition}, pp.\  8895--8904, 2019{\natexlab{c}}.

\bibitem[Liu et~al.(2020{\natexlab{b}})Liu, Hu, Cao, Zhang, and Tong]{closer3d}
Liu, Z., Hu, H., Cao, Y., Zhang, Z., and Tong, X.
\newblock A closer look at local aggregation operators in point cloud analysis.
\newblock In \emph{European Conference on Computer Vision}, pp.\  326--342.
  Springer, 2020{\natexlab{b}}.

\bibitem[Paszke et~al.(2017)Paszke, Gross, Chintala, Chanan, Yang, DeVito, Lin,
  Desmaison, Antiga, and Lerer]{torch}
Paszke, A., Gross, S., Chintala, S., Chanan, G., Yang, E., DeVito, Z., Lin, Z.,
  Desmaison, A., Antiga, L., and Lerer, A.
\newblock Automatic differentiation in pytorch.
\newblock 2017.

\bibitem[Qi et~al.(2017{\natexlab{a}})Qi, Su, Mo, and Guibas]{pointnet}
Qi, C.~R., Su, H., Mo, K., and Guibas, L.~J.
\newblock Pointnet: Deep learning on point sets for 3d classification and
  segmentation.
\newblock In \emph{Proceedings of the IEEE conference on computer vision and
  pattern recognition}, pp.\  652--660, 2017{\natexlab{a}}.

\bibitem[Qi et~al.(2017{\natexlab{b}})Qi, Yi, Su, and Guibas]{pointnet++}
Qi, C.~R., Yi, L., Su, H., and Guibas, L.~J.
\newblock Pointnet++: Deep hierarchical feature learning on point sets in a
  metric space.
\newblock \emph{arXiv preprint arXiv:1706.02413}, 2017{\natexlab{b}}.

\bibitem[Rao et~al.(2020)Rao, Lu, and Zhou]{pointGLR}
Rao, Y., Lu, J., and Zhou, J.
\newblock Global-local bidirectional reasoning for unsupervised representation
  learning of 3d point clouds.
\newblock In \emph{Proceedings of the IEEE/CVF Conference on Computer Vision
  and Pattern Recognition}, pp.\  5376--5385, 2020.

\bibitem[Rubner et~al.(2000)Rubner, Tomasi, and Guibas]{emd}
Rubner, Y., Tomasi, C., and Guibas, L.~J.
\newblock The earth mover's distance as a metric for image retrieval.
\newblock \emph{International journal of computer vision}, 40\penalty0
  (2):\penalty0 99--121, 2000.

\bibitem[Shi et~al.(2020)Shi, Guo, Jiang, Wang, Shi, Wang, and Li]{pvrcnn}
Shi, S., Guo, C., Jiang, L., Wang, Z., Shi, J., Wang, X., and Li, H.
\newblock Pv-rcnn: Point-voxel feature set abstraction for 3d object detection.
\newblock In \emph{Proceedings of the IEEE/CVF Conference on Computer Vision
  and Pattern Recognition}, pp.\  10529--10538, 2020.

\bibitem[Taghanaki et~al.(2020)Taghanaki, Hassani, Jayaraman, Khasahmadi, and
  Custis]{point_mask}
Taghanaki, S.~A., Hassani, K., Jayaraman, P.~K., Khasahmadi, A.~H., and Custis,
  T.
\newblock Pointmask: Towards interpretable and bias-resilient point cloud
  processing.
\newblock \emph{arXiv preprint arXiv:2007.04525}, 2020.

\bibitem[Te et~al.(2018)Te, Hu, Zheng, and Guo]{rgcnn}
Te, G., Hu, W., Zheng, A., and Guo, Z.
\newblock Rgcnn: Regularized graph cnn for point cloud segmentation.
\newblock In \emph{Proceedings of the 26th ACM international conference on
  Multimedia}, pp.\  746--754, 2018.

\bibitem[Thomas et~al.(2019)Thomas, Qi, Deschaud, Marcotegui, Goulette, and
  Guibas]{kpconv}
Thomas, H., Qi, C.~R., Deschaud, J.-E., Marcotegui, B., Goulette, F., and
  Guibas, L.~J.
\newblock Kpconv: Flexible and deformable convolution for point clouds.
\newblock In \emph{Proceedings of the IEEE/CVF International Conference on
  Computer Vision}, pp.\  6411--6420, 2019.

\bibitem[Tsai et~al.(2020)Tsai, Yang, Ho, and Jin]{knnattack}
Tsai, T., Yang, K., Ho, T.-Y., and Jin, Y.
\newblock Robust adversarial objects against deep learning models.
\newblock In \emph{Proceedings of the AAAI Conference on Artificial
  Intelligence}, volume~34, pp.\  954--962, 2020.

\bibitem[Verma et~al.(2019)Verma, Lamb, Beckham, Najafi, Mitliagkas, Lopez-Paz,
  and Bengio]{manifold}
Verma, V., Lamb, A., Beckham, C., Najafi, A., Mitliagkas, I., Lopez-Paz, D.,
  and Bengio, Y.
\newblock Manifold mixup: Better representations by interpolating hidden
  states.
\newblock In \emph{International Conference on Machine Learning}, pp.\
  6438--6447. PMLR, 2019.

\bibitem[Walawalkar et~al.(2020)Walawalkar, Shen, Liu, and
  Savvides]{walawalkar2020attentive}
Walawalkar, D., Shen, Z., Liu, Z., and Savvides, M.
\newblock Attentive cutmix: An enhanced data augmentation approach for deep
  learning based image classification.
\newblock In \emph{ICASSP 2020-2020 IEEE International Conference on Acoustics,
  Speech and Signal Processing (ICASSP)}, pp.\  3642--3646. IEEE, 2020.

\bibitem[Wang et~al.(2018)Wang, Samari, and Siddiqi]{localgcn}
Wang, C., Samari, B., and Siddiqi, K.
\newblock Local spectral graph convolution for point set feature learning.
\newblock In \emph{Proceedings of the European conference on computer vision
  (ECCV)}, pp.\  52--66, 2018.

\bibitem[Wang et~al.(2019)Wang, Sun, Liu, Sarma, Bronstein, and Solomon]{dgcnn}
Wang, Y., Sun, Y., Liu, Z., Sarma, S.~E., Bronstein, M.~M., and Solomon, J.~M.
\newblock Dynamic graph cnn for learning on point clouds.
\newblock \emph{Acm Transactions On Graphics (tog)}, 38\penalty0 (5):\penalty0
  1--12, 2019.

\bibitem[Wu et~al.(2015)Wu, Song, Khosla, Yu, Zhang, Tang, and
  Xiao]{modelnet40}
Wu, Z., Song, S., Khosla, A., Yu, F., Zhang, L., Tang, X., and Xiao, J.
\newblock 3d shapenets: A deep representation for volumetric shapes.
\newblock In \emph{Proceedings of the IEEE conference on computer vision and
  pattern recognition}, pp.\  1912--1920, 2015.

\bibitem[Wu et~al.(2020)Wu, Duan, Wang, Fan, and Guibas]{ifdefense}
Wu, Z., Duan, Y., Wang, H., Fan, Q., and Guibas, L.~J.
\newblock If-defense: 3d adversarial point cloud defense via implicit function
  based restoration.
\newblock \emph{arXiv preprint arXiv:2010.05272}, 2020.

\bibitem[Xiang et~al.(2019)Xiang, Qi, and Li]{generatingadpoint}
Xiang, C., Qi, C.~R., and Li, B.
\newblock Generating 3d adversarial point clouds.
\newblock In \emph{Proceedings of the IEEE/CVF Conference on Computer Vision
  and Pattern Recognition}, pp.\  9136--9144, 2019.

\bibitem[Xu \& Lee(2020)Xu and Lee]{weaklypoints}
Xu, X. and Lee, G.~H.
\newblock Weakly supervised semantic point cloud segmentation: Towards 10x
  fewer labels.
\newblock In \emph{Proceedings of the IEEE/CVF Conference on Computer Vision
  and Pattern Recognition}, pp.\  13706--13715, 2020.

\bibitem[Yan et~al.(2020)Yan, Zheng, Li, Wang, and Cui]{pointasnl}
Yan, X., Zheng, C., Li, Z., Wang, S., and Cui, S.
\newblock Pointasnl: Robust point clouds processing using nonlocal neural
  networks with adaptive sampling.
\newblock In \emph{Proceedings of the IEEE/CVF Conference on Computer Vision
  and Pattern Recognition}, pp.\  5589--5598, 2020.

\bibitem[Yang et~al.(2019{\natexlab{a}})Yang, Zhang, Fang, Ni, Liu, and
  Tian]{adversarialpl}
Yang, J., Zhang, Q., Fang, R., Ni, B., Liu, J., and Tian, Q.
\newblock Adversarial attack and defense on point sets.
\newblock \emph{arXiv preprint arXiv:1902.10899}, 2019{\natexlab{a}}.

\bibitem[Yang et~al.(2019{\natexlab{b}})Yang, Zhang, Ni, Li, Liu, Zhou, and
  Tian]{modelingpl}
Yang, J., Zhang, Q., Ni, B., Li, L., Liu, J., Zhou, M., and Tian, Q.
\newblock Modeling point clouds with self-attention and gumbel subset sampling.
\newblock In \emph{Proceedings of the IEEE/CVF Conference on Computer Vision
  and Pattern Recognition}, pp.\  3323--3332, 2019{\natexlab{b}}.

\bibitem[Yi et~al.(2016)Yi, Kim, Ceylan, Shen, Yan, Su, Lu, Huang, Sheffer, and
  Guibas]{shapepartseg}
Yi, L., Kim, V.~G., Ceylan, D., Shen, I.-C., Yan, M., Su, H., Lu, C., Huang,
  Q., Sheffer, A., and Guibas, L.
\newblock A scalable active framework for region annotation in 3d shape
  collections.
\newblock \emph{ACM Transactions on Graphics (ToG)}, 35\penalty0 (6):\penalty0
  1--12, 2016.

\bibitem[Yun et~al.(2019)Yun, Han, Oh, Chun, Choe, and Yoo]{cutmix}
Yun, S., Han, D., Oh, S.~J., Chun, S., Choe, J., and Yoo, Y.
\newblock Cutmix: Regularization strategy to train strong classifiers with
  localizable features.
\newblock In \emph{Proceedings of the IEEE/CVF International Conference on
  Computer Vision}, pp.\  6023--6032, 2019.

\bibitem[Zhang et~al.(2018)Zhang, Cisse, Dauphin, and Lopez-Paz]{mixup}
Zhang, H., Cisse, M., Dauphin, Y.~N., and Lopez-Paz, D.
\newblock mixup: Beyond empirical risk minimization.
\newblock In \emph{ICLR}. OpenReview.net, 2018.

\bibitem[Zhao et~al.(2019)Zhao, Jiang, Fu, and Jia]{pointweb}
Zhao, H., Jiang, L., Fu, C.-W., and Jia, J.
\newblock Pointweb: Enhancing local neighborhood features for point cloud
  processing.
\newblock In \emph{Proceedings of the IEEE/CVF Conference on Computer Vision
  and Pattern Recognition}, pp.\  5565--5573, 2019.

\bibitem[Zhao et~al.(2020)Zhao, Wu, Chen, and Lim]{isometry3d}
Zhao, Y., Wu, Y., Chen, C., and Lim, A.
\newblock On isometry robustness of deep 3d point cloud models under
  adversarial attacks.
\newblock In \emph{Proceedings of the IEEE/CVF Conference on Computer Vision
  and Pattern Recognition}, pp.\  1201--1210, 2020.

\bibitem[Zheng et~al.(2019)Zheng, Chen, Yuan, Li, and
  Ren]{pointcloudsaliencymaps}
Zheng, T., Chen, C., Yuan, J., Li, B., and Ren, K.
\newblock Pointcloud saliency maps.
\newblock In \emph{Proceedings of the IEEE/CVF International Conference on
  Computer Vision}, pp.\  1598--1606, 2019.

\bibitem[Zhou et~al.(2019)Zhou, Chen, Zhang, Fang, Zhou, and Yu]{dupnet}
Zhou, H., Chen, K., Zhang, W., Fang, H., Zhou, W., and Yu, N.
\newblock Dup-net: Denoiser and upsampler network for 3d adversarial point
  clouds defense.
\newblock In \emph{Proceedings of the IEEE/CVF International Conference on
  Computer Vision}, pp.\  1961--1970, 2019.

\bibitem[Zhou et~al.(2020)Zhou, Chen, Liao, Chen, Dong, Liu, Zhang, Hua, and
  Yu]{lggan}
Zhou, H., Chen, D., Liao, J., Chen, K., Dong, X., Liu, K., Zhang, W., Hua, G.,
  and Yu, N.
\newblock Lg-gan: Label guided adversarial network for flexible targeted attack
  of point cloud based deep networks.
\newblock In \emph{Proceedings of the IEEE/CVF Conference on Computer Vision
  and Pattern Recognition}, pp.\  10356--10365, 2020.

\end{thebibliography}
\bibliographystyle{icml2019}

\end{document}